\documentclass[journal]{IEEEtran}
 
\usepackage{xcolor,soul,framed} 

\colorlet{shadecolor}{yellow}

\usepackage{mathtools}
\usepackage{mathabx}
\usepackage{graphicx} 
\usepackage{multirow}
\usepackage{tabularx}
\usepackage[normalem]{ulem}

\usepackage{cite}
\usepackage{url}
\newcolumntype{Y}{>{\centering\arraybackslash}X}
\ifCLASSINFOpdf

\else
    
\fi

\begin{document}

\title{Classification of Hand Movements from EEG using a Deep Attention-based LSTM Network}

\author{Guangyi~Zhang, ~\IEEEmembership{Student Member,~IEEE}
        Vandad~Davoodnia, ~\IEEEmembership{Student Member,~IEEE,}
        Alireza~Sepas-Moghaddam, 
         Yaoxue~Zhang, ~\IEEEmembership{Senior Member,~IEEE,}
        and~Ali~Etemad,~\IEEEmembership{Member,~IEEE}
\thanks{G. Zhang, V. Davoodnia, A. Sepas-Moghaddam and A. Etemad are with the Department of Electrical and Computer Engineering, Queen's University, Kingston, ON, Canada (e-mail: guangyi.zhang@queensu.ca,
vandad.davoodnia@queensu.ca, alireza.sepasmoghaddam@queensu.ca 
ali.etemad@queensu.ca).}
\thanks{Y. Zhang is with the Department of Computer Science and Technology, Tsinghua University, Beijing, China. (e-mail: zyx@mail.tsinghua.edu.cn).}
}

\maketitle

\begin{abstract}
Classifying limb movements using brain activity is an important task in Brain-computer Interfaces (BCI) that has been successfully used in multiple application domains, ranging from human-computer interaction to medical and biomedical applications. This paper proposes a novel solution for classification of left/right hand movement by exploiting a Long Short-Term Memory (LSTM) network with attention mechanism to learn the electroencephalogram (EEG) time-series information. To this end, a wide range of time and frequency domain features are extracted from the EEG signals and used to train an LSTM network to perform the classification task. We conduct extensive experiments with the EEG Movement dataset and show that our proposed solution our method achieves improvements over several benchmarks and state-of-the-art methods in both intra-subject and cross-subject validation schemes. Moreover, we utilize the proposed framework to analyze the information as received by the sensors and monitor the activated regions of the brain by tracking EEG topography throughout the experiments.
\end{abstract}


\IEEEpeerreviewmaketitle

\section{Introduction} \label{sec:sec1}
Electroencephalogram (EEG) records electrical signals from the brain, thus providing the ability to extract valuable information regarding brain activity. EEG-based Brain-Computer Interfaces (BCI) have been widely used in medical and biomedical applications such as analyzing mental workload and fatigue \cite{kathner2014effects}, diagnosing brain tumors \cite{abdulkader2015brain}, and rehabilitation of central nervous system disorders \cite{daly2008brain}. BCI can also help communicate brain commands and enable the control of artificial limbs \cite{wolpaw2002brain}, especially for people suffering from amyotrophic lateral sclerosis brainstem stroke, brain or spinal injury, cerebreal palsy, muscular dystrophies, and other diseases impairing the control and feedback system between brain and muscles.

In recent years, EEG-based movement analysis and classification have been widely used in various applications, ranging from clinical applications to brain-machine interface and robotics. For example, stroke patients are often asked to make several body movements in response to various visual or electrical stimuli, which allows researchers to monitor the progress of the recovery of the patient's brain injury by analyzing EEG signals \cite{takahashi2012event}. Additionally, such technologies allow for patients with disabilities to control the movements of artificial limbs or exoskeletons. In particular, in order to perform everyday tasks, the control of hand movements is of critical importance for patients \cite{hochberg2006neuronal}. 

To tackle the problem of BCI for hand-movement control, a number of solutions have been proposed in the literature \cite{pfurtscheller2000current, cantillo2014approach, lotte2009comparison}. Generally, two approaches can be used for development of automated methods for BCI, including hand-movement classification from EEG. In the first approach, the system is trained and/or calibrated on the intended user and then used for BCI applications for that same user (intra-subject). While this approach is effective, it does not result in a generalized off-the-shelf solution for a population of patients. The second approach is to develop a generalized solution that performs across subjects once trained with a dataset (cross-subject). While the latter approach is more desired and convenient, it tends to show lower accuracies, typically below the standards required to employ such systems in real products and solutions.

In this paper, a deep learning solution for Left/Right (L/R) hand movement classification using an LSTM network with attention mechanism is proposed. Our proposed method includes three main steps: \textit{i}) data pre-processing is performed to reduce the negative effects of signal artifacts, including cross-talk, noise, and power-line interference; \textit{ii}) time and frequency domain features are extracted from EEG, to then be used as inputs to the LSTM input layer; and \textit{iii}) an attention-based LSTM network is designed to learn the importance of EEG information varying through time, where discriminative information with higher importance are assigned higher scores to better contribute to the classification performance. The architecture of our proposed solution exploits both long and short-term dependencies within the feature manifold. To evaluate the performance and robustness of our solution, we utilize both intra-subject and cross-subject validation schemes in L/R classification experiments. The experimental results are compared with a number of benchmarks as well as the state-of-the-art.


\textbf{Our contributions} are as follows: \textit{i}) The proposed deep model, which has been trained over all the available data ($103$ subjects) using a $10$-fold cross-subject validation scheme, significantly outperforms the state-of-the-art solutions for hand movement classification. \textit{ii}) In order to compare our work to previous studies utilizing the same dataset, we also perform intra-subject classification (using the same network) for each of the $103$ subjects separately, and achieve very high accuracies, outperforming previous studies. \textit{iii}) Lastly, we perform a detailed analysis of brain activity through the different stages of stimuli perception and hand movement, and demonstrate that EEG information flow through the senor pairs are in correspondence with the known and expected neurological function of the brain.

\section{Related Work} \label{sec:sec2}
The majority of the related work on hand movement classification has focused on \textit{intra-subject} validation. Employing this approach often stems from the fact that distinguishing between L/R hand movements can be a challenging task due to the highly subject-dependant nature of brain activities in the visual and motor cortex. Several conventional machine learning methods have been employed using this approach, for instance, in \cite{wang2012comprehensive}, an average accuracy of $64.02\%$ was reported using a Common Spatial Patterns (CSP) approach for $10$ subjects. In \cite{eva2015comparison}, an average accuracy of $88.69\%$ was reported using QDA, while a rough set-based classifier was used in \cite{szczuko2016rough} and \cite{szczuko2018comparison}, reporting average accuracies of $60\%$ and $68\%$ respectively. However, the aforementioned traditional classifiers often cannot model the non-linearities observed in high-dimensional multi-channel EEG and feature-sets extracted from the data. This especially becomes an issue when attempting to model cross-subject relationships within the dataset.

Training a single model capable of learning to classify hand movements from EEG and generalizing the learned input-output relationships to the entire dataset (\textit{cross-subject}) is quite challenging. The results presented in \cite{szczuko2017real} showed that the average accuracy dropped from $87\%$ to chance level ($50\%$), when utilizing a cross-subject scheme instead of an intra-subject one, using the proposed Maximum Discernibility Algorithm (MDA). Moreover, a large-scale synchronization analysis using Phase Locking Value (PLV) in \cite{loboda2014discrimination}, defined a criteria that could successfully distinguish between L/R hand movement and obtained an average accuracy of $78.95\%$. Other than conventional machine learning classifiers and statistic analysis, deep learning techniques have been employed for this task. For instance, in one of the few deep learning solutions, a set of time and frequency domain features were extracted and fed to an Artificial Neural Network (ANN) \cite{huong2017classification}, achieving an accuracy of $68\%$.

In addition to the task of hand movement classification, the task of imagery classification focuses on mental activity when imagining left and right hand movements as opposed to physically performing them. In this context, several classical machine learning methods such as Logistic Regression \cite{tomioka2007logistic}, k-Nearest Neighbor (kNN) \cite{bhattacharyya2011performance}, and Na\"ive Bayes (NB) \cite{bhattacharyya2011performance} have been used with intra-subject validation. In addition to classical machine learning methods, several deep learning solutions have also been utilized. For example, in \cite{tang2017single, tabar2016novel}, a Convolutional Neural Network (CNN) was used. An LSTM-based model was proposed in \cite{wang2018lstm}, outperforming the state-of-the-art solutions including methods based on other deep networks. In cross-subject validation, a very high accuracy was achieved using a parallel or cascade combination of CNN and LSTM networks \cite{zhang2017eeg}. It should be mentioned that the imagery EEG classification naturally uses a dataset separate from the movement dataset, and hence, a direct comparison between the two tasks is not valid. However, a review of the methods used for imagery classification can provide further inspiration for new solutions for movement classification. 

Table \ref{tab:related work} summarizes the main works in the area and characteristics of the proposed solutions in terms of method and classification tasks. The table also includes information about the datasets and validation protocols and schemes considered by these methods for performance assessment. The solutions are sorted according to their publication date. For comparison, the characteristics of our method proposed in this paper are also included in Table \ref{tab:related work}. The Table points to two clear areas in the literature that merit additional investigation and research: 
\begin{itemize}
    \item First, most of the works have performed intra-subject validation, while the more challenging task of cross-subject validation (which is also more indicative of generalization) has been largely disregarded.
    \item Second, most of the prior works have utilized classical machine learning models, indicating room for improvement using more advanced deep learning techniques.
\end{itemize}

\newlength{\temp}
\settowidth{\temp}{my caption line xxx }
\newcolumntype{T}{>{\centering\arraybackslash}X}
\begin{table*}
\centering
\caption{\hspace{70pt} The Main Characteristics of the Related Work \newline \scriptsize 1) \textit{M-EEG} denotes physical left/right hand movements, while \textit{IM-EEG} denotes imagery left/right hand movements in the PhysioNet EEG Motor Movement/Imagery Dataset; 2) \textit{BCI Comp} denotes the imagery dataset used in the BCI Competition; 3) \textit{Comp-IV.2a} and \textit{Comp-IV.2b} denote Dataset 2a and Dataset 2b of the BCI Competition IV, respectively; 4) \textit{Comp-II.III} denotes Dataset III in the BCI Competition II.}
\label{tab:related work}
\scriptsize
\begin{tabularx}{\textwidth}{cTTccTTTT}
	\hline
	\multirow{2}{*}{Ref.}                & \multirow{2}{*}{Year} & Method                           & \multirow{2}{*}{Method}    & Feature           & \multirow{2}{*}{Task}       & \multirow{2}{*}{Dataset} & Validation Protocol & Validation \\              &                      & {Type}                           &                            & Selection         &                             &                          & (No. of Subjects)      & Scheme                               \\
	\hline \hline
	\cite{tomioka2007logistic}           & 2007                  & Classical                        & LR                         & No                & Imagery                 & BCI Comp                & Intra-Sub ($29$)       & $50$:$50$ Split                      \\
    \cite{bhattacharyya2011performance}  & 2011                  & Classical                        & KNN, NB                    & Yes               & Imagery                 & BCI Comp-II.III         & Intra-Sub ($1$)        & $50$:$50$ Split                      \\ 
    \cite{wang2012comprehensive}         & 2012                  & Classical                        & CSP                        & No                & Movement                & M-EEG                   & Intra-Sub ($10$)       & $3$-Fold                             \\ 
    \cite{loboda2014discrimination}      & 2014                  & Classical                        & PLV                        & No                & Movement                & M-EEG                   & Cross-Sub ($103$)      & N/A                                  \\   
    \cite{eva2015comparison}             & 2015                  & Classical                        & QDA                        & No                & Movement                & M-EEG                   & Intra-Sub ($103$)      & N/A                                  \\
    \cite{tabar2016novel}                & 2015                  & Deep                             & CNN+SAE                    & No                & Imagery                 & BCI Comp-IV.2b          & Intra-Sub ($9$)        & $10\times10$-Fold                    \\     
    \cite{szczuko2016rough}              & 2016                  & Classical                        & Rought set                 & No                & Movement                & M-EEG                   & Intra-Sub ($106$)      & $50$:$50$ Split                      \\     
    \cite{tang2017single}                & 2017                  & Deep                             & CNN                        & No                & Imagery                 & N/A                     & Intra-Sub ($2$)        & $80$:$20$ Split                      \\
    \cite{zhang2017eeg}                  & 2017                  &Deep                              & CNN+LSTM                   & No                & Imagery                 & IM-EEG                  & Cross-Sub ($108$)      & $75$:$25$ Split                      \\
    \cite{szczuko2017real}               & 2017                  & Classical                        & MDA                        & No                & Movement                & M-EEG                   & Intra-Sub ($106$)      & $65$:$35$ Split                      \\
    \cite{huong2017classification}       & 2017                  & Deep                             & ANN                        & No                & Movement                & M-EEG                   & Cross-Sub ($109$)      & $10$-Fold                            \\
    \cite{szczuko2018comparison}         & 2018                  & Classical                        & Rough set                  & No                & Movement                & M-EEG                   & Intra-Sub ($106$)      & $65$:$35$ Split                      \\
    \cite{wang2018lstm}                  & 2018                  & Deep                             & LSTM                       & No                & Imagery                 & BCI Comp-IV.2a          & Intra-Sub ($9$)        & $5\times5$-Fold                      \\
    \hline
    Ours                                 & 2019                  & Deep                             & LSTM+Attention                   & No                & Movement                & M-EEG                   & Intra-Sub ($103$)      & $10$-Fold                            \\ 
    Ours                                 & 2019                  & Deep                             & LSTM+Attention                    & No                & Movement                & M-EEG                   & Cross-Sub ($103$)      & $10$-Fold                            \\ 
    \hline \hline
\end{tabularx}
\end{table*}

\vspace{-2mm}
\section{Proposed Method}
In this section, we present the proposed method, including the pre-processing steps, feature extraction, and the deep learning solution. In the rest of this paper, the description of notations used is as follow: `$\mathnormal{a}$' represents a scalar, `$\mathbf{a}$' represents a vector, `$\mathbf{A}$' represents a matrix.

\subsection{Pre-processing}
Data pre-processing was performed to reduce the negative effects of signal artifacts, including cross-talk, noise, and power-line interference. In this context, out of the $64$ EEG channels available in the EEG test material, the $10$ central sensors were discarded due to their non-symmetric nature \cite{lin2010eeg}. The utilized sensor pairs were selected based on the topology described in \cite{lin2010eeg}. Then, two filters were applied to the new $27$ differential EEG channels: a notch filter removed $50$ \textit{Hz} power line interference and a band-pass filter was applied to allow a frequency range of $0.5-70$ \textit{Hz} to pass through, thus minimizing artifacts such as noise often present in this frequency range \cite{alomari2013automated}. Normalization of EEG amplitude was then carried out as the last step to minimize the difference in EEG amplitudes using min-max normalization with across different subjects. 

\subsection{Time and Frequency Domain Feature Extraction}
Successive to pre-processing, a number of time and frequency domain features were extracted in order to be used as inputs for the proposed method. EEG is known as a non-stationary time-series signal where nonlinear features are often used for representation and classification tasks. Feature extraction was performed on a $2$-second segments for each trial. The effect of different segment sizes on the performance of hand movement classification task will be studied in Section 5. Time and frequency domain features were subsequently extracted from each time-step. Time-domain features included: \textit{i}) mean, \textit{ii}) variance, \textit{iii}) skewness, \textit{iv}) kurtosis, \textit{v}) zero crossings, \textit{vi}) absolute area under the signal, and \textit{vii}) peak-to-peak distance. Extracted frequency-domain features consisted of relative band power in four frequency bands, notably \textit{i}) delta ($0.5-4$ \textit{Hz}), \textit{ii}) theta ($4-8$ \textit{Hz}), \textit{iii}) alpha ($8-12$ \textit{Hz}), and \textit{iv}) beta ($12-30$ \textit{Hz}). Table \ref{eq_table} presents the mathematical equations for these features, where a total of $297$ features ($27$ channels $\times$ $11$ features per channel) were extracted from each time-step.

\begin{table}
\caption{Time and Frequency Domain Features} 
\label{eq_table}
\begin{tabularx}{1.0\columnwidth}{cY}
	\hline
    {Method} &  {Formula}\\
	\hline
	\hline
	 Mean & {$ \mu = \displaystyle\frac{1}{N}\displaystyle\sum_{i=1}^{N}x_i $} \\
	 Variance & {$ \sigma^2 = \displaystyle\frac{1}{N}\displaystyle\sum_{i=1}^{N}(x_i - \mu)^2 $} \\
	 Skewness & {$ S = \displaystyle\frac{\displaystyle\frac{1}{N}\displaystyle\sum_{i=1}^{N}(x_i -                 \mu)^3}{(\displaystyle\frac{1}{N-1}\displaystyle\sum_{i=1}^{N}(x_i - \mu)^2)^{3/2}} $} \\
	 Kurtosis & {$ K = \displaystyle\frac{\displaystyle\frac{1}{N}\displaystyle\sum_{i=1}^{N}(x_i - \mu)^4}{(\displaystyle\frac{1}{N}\displaystyle\sum_{i=1}^{N}(x_i - \mu)^2)^2} - 3 $} \\
	 Zero-crossing & {$ zc = \displaystyle\sum_{i=1}^{N-1}1_{\displaystyle{\rm I\!R}_{<0}}(x_i x_{i-1}) $} \\
	 Absolute area under signal & {$ simps = \displaystyle\int_{a}^{b}|f(x)|dx $} \\ 
	 Peak to Peak & $\displaystyle pk2pk = max(\mathbf{x})- min(\mathbf{x})$ \\
	 \hline 
	 Amplitude spectrum density & {$ \displaystyle\hat{X}(\omega)=\displaystyle\frac{1}{\sqrt{T}}\displaystyle\int_{0}^{T}x(t)\exp^{-i\omega t}dt $} \\
	 Power spectrum density & $ S_{xx}(\displaystyle\omega)=\displaystyle\lim_{T\to\infty}E\Big[|\displaystyle\hat{X}(\omega)|^{2}\Big] $ \\
	 power of each frequency band & $ P = \displaystyle\frac{1}{\pi}\displaystyle\int_{\omega_1}^{\omega_2}S_{xx}(\omega)d\omega $ \\
	\hline
\end{tabularx}
\end{table}

\vspace{-2mm}

\subsection{Proposed Deep Learning Solution} 
To detect and classify very subtle spatio-temporal changes in our feature space that correspond to the intended movements, a solution capable of remembering and eventually \textit{aggregating} these transitions across the dataset is required. Addressing both these requirements formed the intuition behind our proposed solution of using an LSTM network with the \textit{attention} mechanism. LSTM has been widely utilized for learning and classifying time-series data including bio-signals \cite{tsiouris2018long, zhang2017eeg}. Moreover, recent studies have successfully used LSTM architectures for EEG analysis given the time-dependant nature of these signals \cite{wang2018lstm}. Furthermore, attention-based LSTM has been used in other tasks requiring remembering and aggregation of feature embeddings, notably natural language processing (NLP) \cite{zhou2016attention, wang2016attention}. In the field of NLP, the importance of different words vary depending on context and role in the sentence. Similarly, the importance of information obtained in time-steps of EEG signals are also discrepant and task-dependent. Thus we believe the attention-based LSTM architecture can improve classification performance using EEG signals by focusing on essential task-relevant features from different time-steps. In the following subsection we describe the architecture of an LSTM cell as well as the attention mechanism.

\subsubsection{LSTM Network} RNNs can be used to extract higher dimensional dependencies from sequential data such as EEG time-series \cite{lipton2015critical}. RNN units have connections not only between the subsequent layers, but also among themselves to capture information from previous inputs. Traditional RNNs can easily learn short-term dependencies; however, they have difficulties in learning long-term dynamics due to the vanishing and exploding gradient problems. LSTM is a type of RNN that addresses the vanishing and exploding gradient problems by learning both long- and short-term dependencies \cite{greff2017lstm}. 

\begin{figure}[t]
    \begin{center}
    \includegraphics[width=0.9\columnwidth]{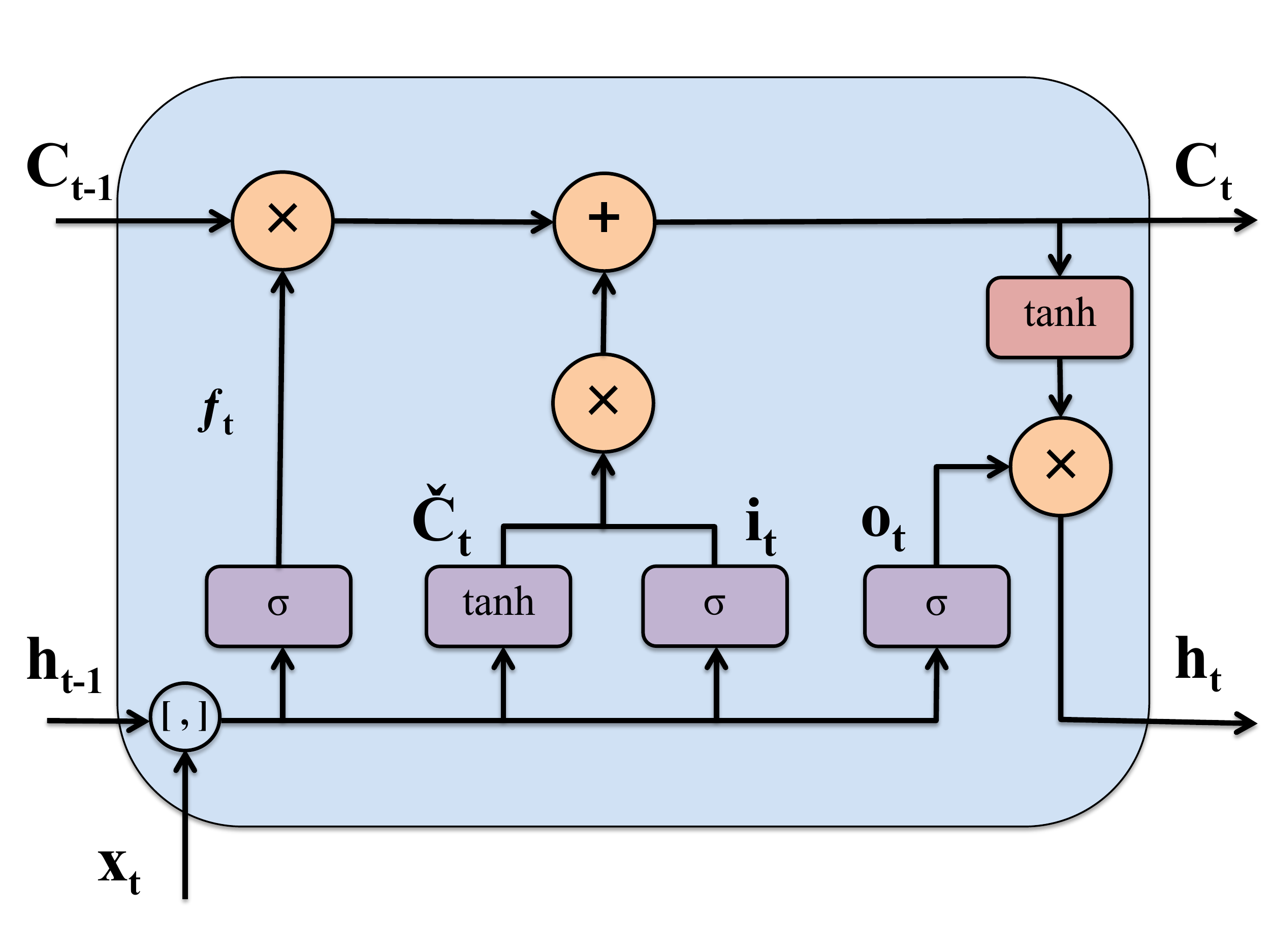}
    \caption{An LSTM cell architecture is illustrated. [,] denotes array concatenation.}\label{fig:LSTM}
    \end{center}
\end{figure}

An LSTM network is composed of cells, whose outputs evolve through the network based on past memory content. The cells have a common cell state, keeping long-term dependencies along the entire LSTM chain of cells. The flow of information is then controlled by the input gate ($\mathbf{i_{t}}$) and forget gate ($\mathbf{f_{t}}$), thus allowing the network to decide whether to forget the previous state ($\mathbf{C_{t-1}}$) or update the current state ($\mathbf{C_t}$) with new information. The output of each cell (hidden state) is controlled by an output gate ($\mathbf{o_{t}}$), allowing the cell to compute its output given the updated cell state (see Figure \ref{fig:LSTM}).

The formulas describing an LSTM cell architecture are presented as:
\begin{equation}
\mathbf{i_{t}} = \sigma(\mathbf{W_{i}}\cdot[\mathbf{h_{t-1}}, \mathbf{x_{t}}] + \mathbf{b_{i}}),
\end{equation}\vspace{-5mm}
\begin{equation}
\mathbf{f_{t}} = \sigma(\mathbf{W_{f}}\cdot[\mathbf{h_{t-1}}, \mathbf{x_{t}}] + \mathbf{b_{f}}),
\end{equation}\vspace{-5mm}
\begin{equation}
\mathbf{C_{t}} = \mathbf{f_t}*\mathbf{C_{t-1}} + \mathbf{i_t} * tanh(\mathbf{W_c} \cdot[\mathbf{h_{t-1}}, \mathbf{x_{t}}]+\mathbf{b_c}) ,
\end{equation}\vspace{-5mm}
\begin{equation}
\mathbf{o_{t}} = \sigma(\mathbf{W_{o}}\cdot[\mathbf{h_{t-1}}, \mathbf{x_{t}}] + \mathbf{b_{o}}), 
\end{equation}\vspace{-5mm}
\begin{equation}
\mathbf{h_{t}} = \mathbf{o_{t}} * tanh(\mathbf{C_t}) ,
\end{equation}
where $\sigma (x) = \frac{1}{1+e^{-x}}, tanh(x) = \frac{2}{1+e^{-2x}} - 1$, $\mathbf{h_t}$ is the hidden state at time step \textit{t}, $\mathbf{C_{t-1}}$ is the cell state at time step \textit{t}, $\mathbf{x_{t}}$ is the input features fed to the cell, $\mathbf{W_{f}},\mathbf{W_{i}},\mathbf{W_{c}},\mathbf{W_{o}}$ are the weights, and $\mathbf{b_{f}},\mathbf{b_{i}},\mathbf{b_{c}},\mathbf{b_{o}}$ are the biases that can be obtained by backpropagation through time.

\subsubsection{Attention Mechanism} 
Attention mechanism can improve the performance of LSTM by focusing on certain time-steps with the most discriminative information. Unlike conventional LSTM networks that use their last hidden state as output, an LSTM network with attention mechanism multiplies the output hidden states by trainable weights, as shown in Figure \ref{fig:1-5}, thus capturing more discriminative task-related features. This can be formulated as:
\begin{equation}
\mathbf{h_i} = LSTM(\mathbf{s_{i}}), i\in [1,L],
\end{equation}
where, $\mathbf{h_i}$ is the output hidden state vector for the $i^{th}$ LSTM cell corresponding to the $i^{th}$ input, and $L$ is the number of cells in each recurrent layer of the LSTM network. To capture the importance of each hidden state, attention mechanism is defined as follows:
\begin{equation}
\mathbf{u_i} = tanh(\mathbf{W_s}\mathbf{h_i} + \mathbf{b_s}),
\end{equation}\vspace{-2mm}
\begin{equation}
\mathbf{\alpha_i} = \frac{exp(\mathbf{u_i)}}{\sum_j{exp(\mathbf{u_j})}},
\end{equation}\vspace{-2mm}
\begin{equation}
\mathbf{v} = \sum_i{\mathbf{\alpha_i} \mathbf{h_i}},
\end{equation}
where vector $\mathbf{v}$ is the attention layer's output, and $\mathbf{W_{s}}$ and $\mathbf{b_s}$ are trainable parameters.

\subsubsection{Proposed Network}
\begin{figure}[t]
    \begin{center}
    \includegraphics[width=0.9\columnwidth]{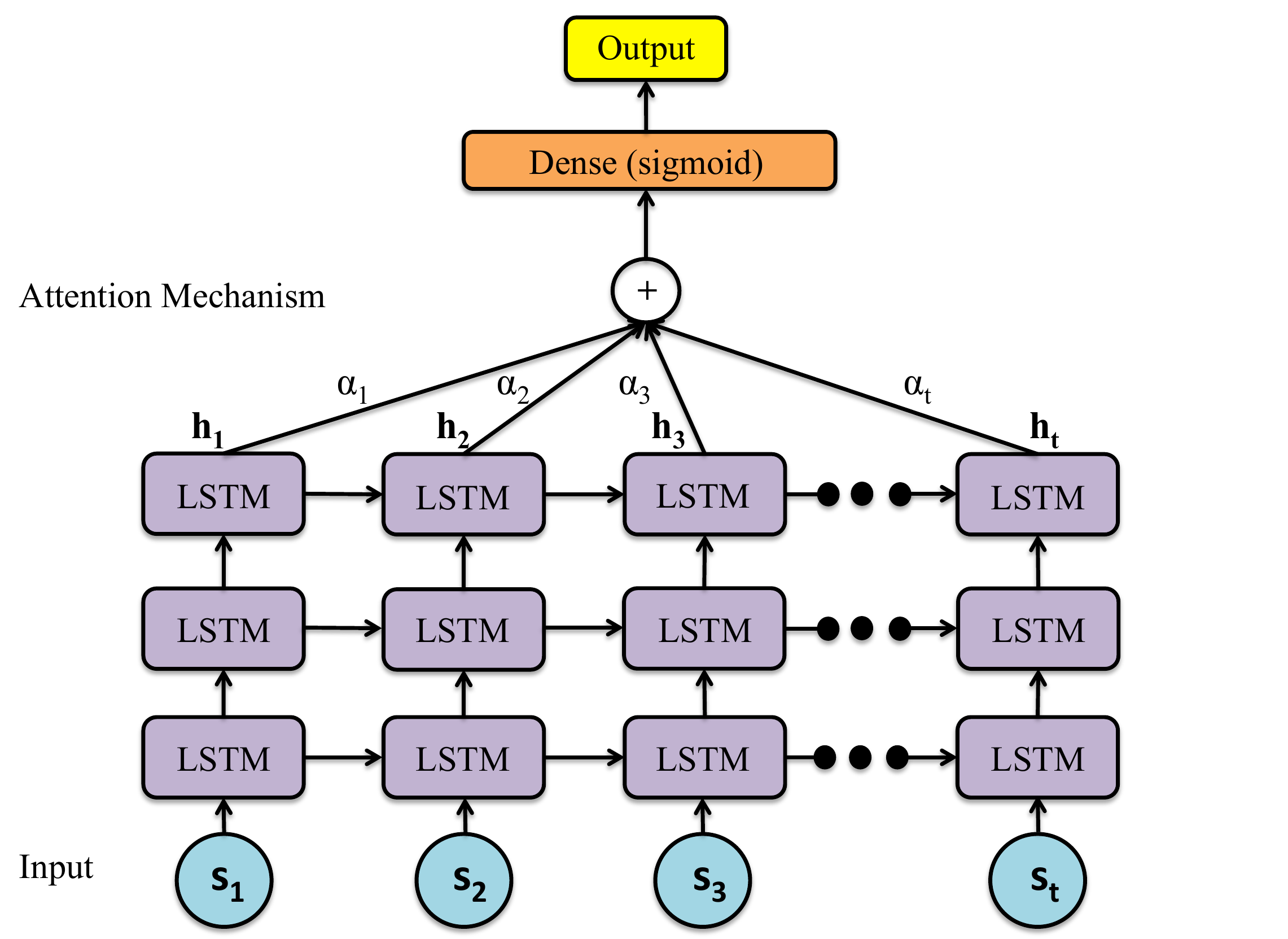} 
    \caption{The overview of our proposed LSTM+attention solution is presented.}
    \label{fig:1-5}
    \end{center}
\end{figure}

All the $297$ features from the $7$ time-steps in each segment (as described earlier), were fed to $7$ individual cells of the first LSTM layer. We employed three stacked $7$-cell layers in our network. The final LSTM layer was followed by an attention layer, which was in turn followed by a fully connected layer with a sigmoid activation function to predict the probability of each class. This network The architecture is depicted in Figure \ref{fig:1-5}.

\section{Experiment Setup and Evaluation}
This section presents the test material, optimal LSTM hyper-parameter setting, validation protocols, and the state-of-the-art and benchmark recognition solutions considered for comparison purposes.

\subsection{Dataset}
\begin{figure}[t!]
    \begin{center}
    \includegraphics[width=0.9\columnwidth]{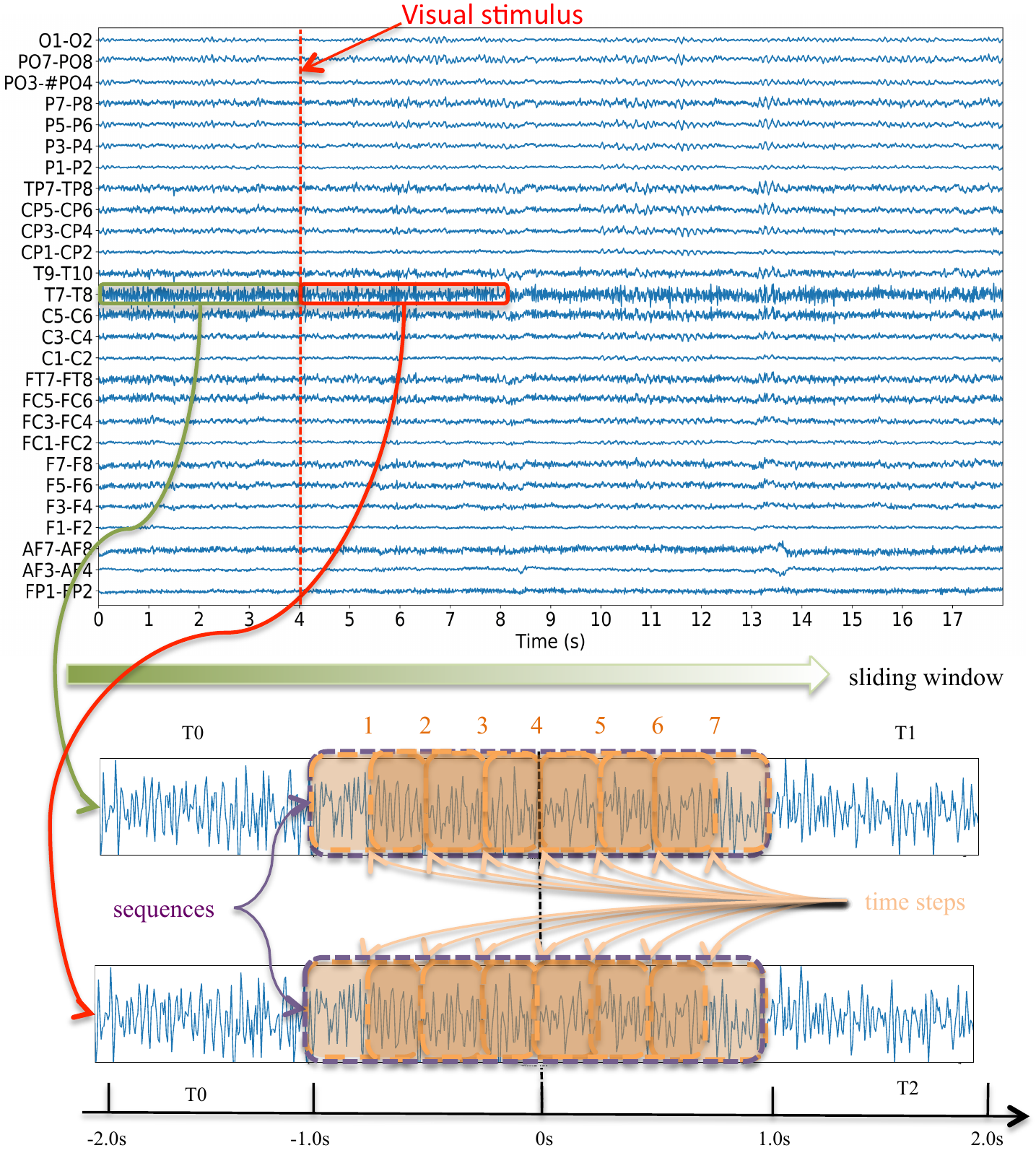} 
    \caption{An overview of the EEG data, the movement segments (2-second long LSTM sequence consists of 7 time steps with 50\% overlap between adjacent windows), and the sliding window used during training/classification is illustrated.}
    \label{fig:1-2}
    \end{center}
\end{figure}

The EEG Movement Dataset\footnote{https://physionet.org/pn4/eegmmidb/}$^,$\footnote{www.bci2000.org} \cite{schalk2004bci2000, physiotoolkitphysionet} was used in this study. The dataset includes $109$ subjects and has been collected using a BCI $2000$ system. Participants were asked to perform three actions: rest ($T_0$), left hand movement ($T_1$), and right hand movement ($T_2$). Each experiment consisted of $15$ iterations, where $T_0$ was followed by a visual stimulus, randomly selecting either $T_1$ or $T_2$. This $15$-pair movement process was repeated $3$ times. Accordingly, the dataset contains a total of $103$ subjects $\times$ $3$ experiments $\times$ $15$ movements, for a total of $4635$ movements. The dataset contains $64$-channels of EEG, recorded at a sampling frequency of $160$ \textit{Hz}. Figure \ref{fig:1-2} illustrates a sample EEG recording and the three actions $T_0$, $T_1$, and $T_2$. 

Out of the $109$ subjects in the dataset, the data from $6$ particular subjects ($43$, $88$, $89$, $92$, $100$, and $104$) had low signal to noise ratio. Therefore, they were removed from our dataset. Rejection of poor-quality samples has been performed in the literature for the same dataset \cite{loboda2014discrimination, eva2015comparison, szczuko2016rough, szczuko2018comparison, szczuko2017real}.

\subsection{LSTM Hyper-Parameters Setting} 
A number of hyper-parameters for the network were explored and tuned to achieve the best results for our proposed model. Notably, these hyper-parameters include: recurrent depth, batch size, number of training epochs, LSTM hidden layer size, dropout rates applied after input layer and the following three stacked LSTM layers ($D_0, D_1, D_2, D_3 $), and the weight matrix $L2$ regularization coefficient of each LSTM layer. Additionally, some hyper-parameters were tuned for the stochastic Adam optimizer \cite{kingma2014adam}, such as learning rate ($l_r$) and exponential decay rates for the first and second movement estimates ($\beta_1$ and $\beta_2$). The optimum values for these parameters are presented in Table \ref{eq_hyper}. A different set of parameters was assigned for each validation scheme (cross-subject and intra-subject) to maximize performance. A binary cross-entropy loss function $L = -y\log(p)+(1-y)\log(1-p)$ was employed for training.

vspace{-2mm}

\begin{table}[t]
\begin{center}
\centering
\caption{Training Hyper-Parameters}\label{eq_hyper}
\footnotesize 
\scalebox{1} {
\begin{tabularx}{\columnwidth}{cYY}
	\hline
    {Hyperparamters} &  {Cross-Subject} & {Intra-Subject}\\
    \hline
	Recurrent depth               & $3$   & $3$ \\ 
	Batch size                    & $32$  & $2$\\ 
	Number of training epochs     & $100$ & $10$ \\ 
	LSTM hidden layer size        & $256$ & $256$ \\ 
    \multirow{4}{*}{Dropout rates}                 & $D0=0.0$ & $D0=0.7$ \\
                                & $D1=0.2$ & $D1=0.2$ \\
                                & $D2=0.1$ & $D2=0.1$ \\
                                & $D3=0.2$ & $D3=0.1$ \\ 
    \hline
	\multirow{2}{*}{Learning}   & \multicolumn{2}{c}{$L2=0.001$, $l_r = 0.001$} \\  & \multicolumn{2}{c}{$\beta_1 = 0.9$, $\beta_2 = 0.999$} \\
	\hline
\end{tabularx}
}
\end{center}
\end{table}

\subsection{Validation Protocols and Benchmarking}
To rigorously evaluate the performance for our proposed solution, we utilized both intra-subject and cross-subject validation schemes. Both schemes used $10$-fold cross-validation, where no overlap existed in the training and testing segments at each fold, as previous studies with such overlaps have shown to yield very high accuracies as expected \cite{zhang2017eeg}. True Positive (TP), False Negative (FN), False Positive (FP), and True Negative (TN) were used to calculate the performance metrics, namely Precision, Recall, and Accuracy, which are formulated as follows: 
\begin{equation}
Precision = \frac{TP}{TP + FP},
\end{equation}
\begin{equation}
Recall = \frac{TP}{TP + FN},
\end{equation}
\begin{equation}
Accuracy = \frac{TP + TN}{TP + TN +FP +FN}.
\end{equation}

\subsubsection{Solutions for Cross-Subject Scheme}
Solutions from related works include: PLV \cite{loboda2014discrimination} and ANN \cite{huong2017classification}, which have been discussed in Section \ref{sec:sec2}. As studies performing cross-subject validation on this challenging dataset were not very common, we implemented a number of other techniques including the classical machine learning and deep learning solutions that have been successfully implemented in the imagery task (as mentioned in the related work) to better compare with our proposed solution. The conventional benchmarks include SVM \cite{bhattacharyya2011performance}, Na\"ive Bayes \cite{bhattacharyya2011performance}, decision tree \cite{aydemir2014decision}, logistic regression \cite{tomioka2007logistic}, and random forest \cite{bentlemsan2014random}. The SVM used an $8^{th}$ degree polynomial kernel and the random forest used $30$ estimators up to a depth of $2$. These parameters were tuned empirically in order to achieve the best results. The benchmarking solutions included deep learning methods as well. First, we used a $3-$layer $2$D-CNN, accepting a $2$D matrix of $297$ features as inputs. The network had a kernel size of $3\times3$, and feature maps of $32$, $64$, and $128$, respectively for the first, second, and third convolutional layers. A VGG-$16$ CNN was also used for benchmarking. This model was pre-trained on ImageNet \cite{deng2009imagenet} and fine-tuned for our application. This model was presented with inputs in the form of $3$D matrices, which were achieved by re-sizing the feature matrices to $180\times150\times1$ using linear interpolation. The output of the VGG-$16$ convolution layers was followed by $2$ fully connected layers that used ReLu activation and a final output layer with sigmoid activation for estimating the class probabilities. Lastly, an LSTM network without the attention mechanism was also used for benchmarking. In this model, similar hyperparameters as our proposed attention-based method were used (see Table \ref{eq_hyper}).

\begin{table*}[t!]
\caption{Effect of Segment Size} \label{tab:window}
\centering
\small
\begin{tabular}{ccccccccc}
	\hline
    {Size (seconds)} &            {$0.25$} & {$0.5$} & {$0.75$} & {$1.0$} & {$1.25$} & {$1.5$} & {$1.75$} & {$2.0$} \\
    \hline
    {Accuracy $\pm$ SD} & {$82.18\pm 2.1$} & {$80.23\pm 1.8$} & {$79.1\pm1.5$} & {$80.3 \pm 1.9$} & {$80.8\pm 1.7$} & {$81.1\pm 0.9$} & {$81.2\pm 2.0$} & {$83.2\pm 1.2$} \\
	\hline
\end{tabular}
\end{table*}

\vspace{2mm}

\subsubsection{Solutions for Intra-Subject Scheme}
As discussed earlier, the main goal for this work is to tackle the more challenging task of cross-subject generalization. However, to further test our method and compare to the state-of-the-art, we compared our proposed method with solutions from previous studies namely CSP \cite{wang2012comprehensive}, QDA \cite{eva2015comparison}, rough set-based \cite{szczuko2016rough, szczuko2018comparison}, and MDA \cite{szczuko2017real} described in Section \ref{sec:sec2}. We did not attempt to utilize more benchmarking solutions in this scheme, as most of the related works in this area utilize intra-subject validation, thus comparing to those works was deemed sufficient.

\section{Results and Discussion}
In this section, we report the conducted experiments and performance. First, we study the effect of segment size. Then, we demonstrate the performance of our proposed method along with comparisons to other machine learning techniques and previous studies. Next, we perform a feature analysis, and finally, we discuss the most dominant sensors when maximum accuracy is achieved using our method, followed by an analysis of the flow of information during the experiments.

\textbf{Effect of Segment Size:} In order to select the optimum segment size for feature extraction, we experimented with different segment sizes ($0.25, 0.5, 0.75, 1.0, 1.25, 1.5, 1.75$, and $2.0$ seconds), and evaluated the performance of the model with cross-subject validation. Table \ref{tab:window} presents that the highest classification accuracy and minimum standard deviation with $10$-fold cross-validation were achieved when the segment size was $2$ seconds long. As a result, in this study, the segment size was set to $2$ seconds for feature extraction and classification.

\textbf{Performance:} 
Table \ref{tab:t1} shows the average accuracy, precision, and recall along with their standard deviations for the proposed method and other benchmarking solutions in the cross-subject scheme setting. These results demonstrate the robust performance of our proposed model compared to other methods. The results show that our proposed model significantly outperforms the best performing benchmark, i.e., PLV, by a considerable $5\%$ accuracy. Additionally, Table \ref{t2} reports the accuracy, precision, and recall rates obtained for the intra-subject scheme, showing near-perfect performance, while the previous work with the best performance achieved an accuracy of $88.6\%$ \cite{eva2015comparison}.

\begin{table}[t]
\centering
\caption{Comparison of Different Methods using Cross-subject Scheme} \label{tab:t1}
\footnotesize
\begin{tabular}{cccc}
	\hline
	Methods & Accuracy $\pm$ SD & Precision $\pm$ SD & Recall $\pm$ SD \\
	\hline
	\hline
	PLV \cite{loboda2014discrimination} & $78.9$ & $-$ & $-$\\
	ANN \cite{huong2017classification} & $68.0$ & $-$ & $-$\\
	\hline
	SVM & $62.4 \pm 2.1$ & $61.5 \pm 1.7$ & $62.4 \pm 2.1$\\
	Logistic Regression & $52.9 \pm 1.4$ & $52.4 \pm 2.1$ & $51.1 \pm 1.3$\\
	Decision Tree & $51.0 \pm 1.3$ & $50.3 \pm 1.3$ & $50.3 \pm 1.3$\\
	Random Forest & $53.0 \pm 1.2$ & $52.9 \pm 1.6$ & $61.5 \pm 1.4$\\
	Naive Bayes & $51.1 \pm 1.3$ & $50.7 \pm 1.2$ & $51.2 \pm 1.9$\\
	\hline
	3-layer 2D-CNN & $63.2\pm1.3 $ & $63.1 \pm 1.5$ & $63.1 \pm 1.2$\\
	VGG-16 & $53.2 \pm 1.3$ & $53.1 \pm 2.1$ & $53.2 \pm 1.3$\\
	LSTM & $77.2 \pm 2.5$ & $77.2 \pm1.5$ & $76.9 \pm 1.1$\\
	\hline
	\textbf{LSTM + Attention} & $\mathbf{83.2 \pm 1.2}$ & $\mathbf{83.7 \pm 1.2}$ & $\mathbf{82.2 \pm 2.1}$\\
	\hline
\end{tabular}
\end{table}

\begin{table}[t]
\caption{Comparison of Different Methods using Intra-subject Scheme} 
\label{t2}
\centering
\begin{tabular}{cccc}
	\hline
    {Method} & {Accuracy $\pm$ SD} & {Precision $\pm$ SD} & {Recall $\pm$ SD}\\
	\hline
	\hline
	 CSP \cite{wang2012comprehensive} & {$64.0$} & $-$ & $-$\\
	 QDA \cite{eva2015comparison} & $88.6$ & $-$ & $-$ \\
     Rough set \cite{szczuko2016rough} & {$60.0$} & $-$ & $-$ \\
     Rough set \cite{szczuko2018comparison} & {$68.0$} & $-$ & $-$ \\
     MDA \cite{szczuko2017real} & {$87.0$} & $-$ & $-$ \\
	\hline
    {\textbf{LSTM + Attention}}  & {$\mathbf{98.3 \pm 0.9}$} & {$\mathbf{94.7 \pm 1.1}$} & {$\mathbf{95.9 \pm 1.7}$} \\
	\hline
\end{tabular}
\end{table}

\begin{figure}[t]
    \begin{center}
    \includegraphics[width=1.0\columnwidth]{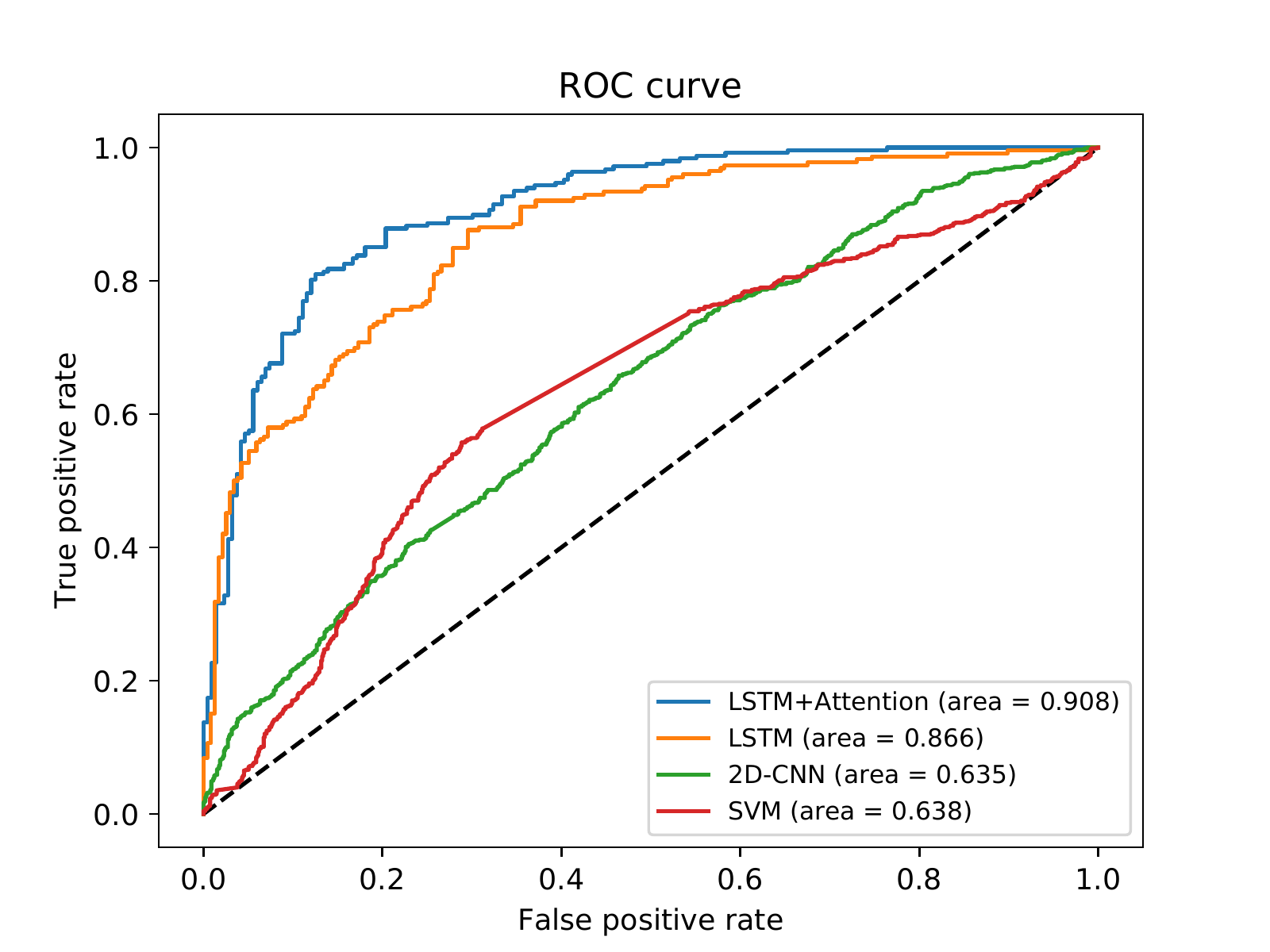} 
    \caption{The ROC curves and corresponding AUCs are presented for our proposed LSTM network and the top-$3$ benchmarking solutions.}
    \label{fig:5-4}
    \end{center}
\end{figure}

The Receiver Operating Characteristic (ROC) curve, which shows the changes of the TP rate with respect to the FP rate, for the proposed model and the top three benchmarking solutions (in the cross-subject scheme) is illustrated in Figure \ref{fig:5-4}. 
This figure also includes the Area Under the Curve (AUC) values that reveal the superiority of our proposed approach over the top three benchmarks with an AUC of $0.908$.

\begin{figure}[t!]
    \begin{center}
    \includegraphics[width=1.0\columnwidth]{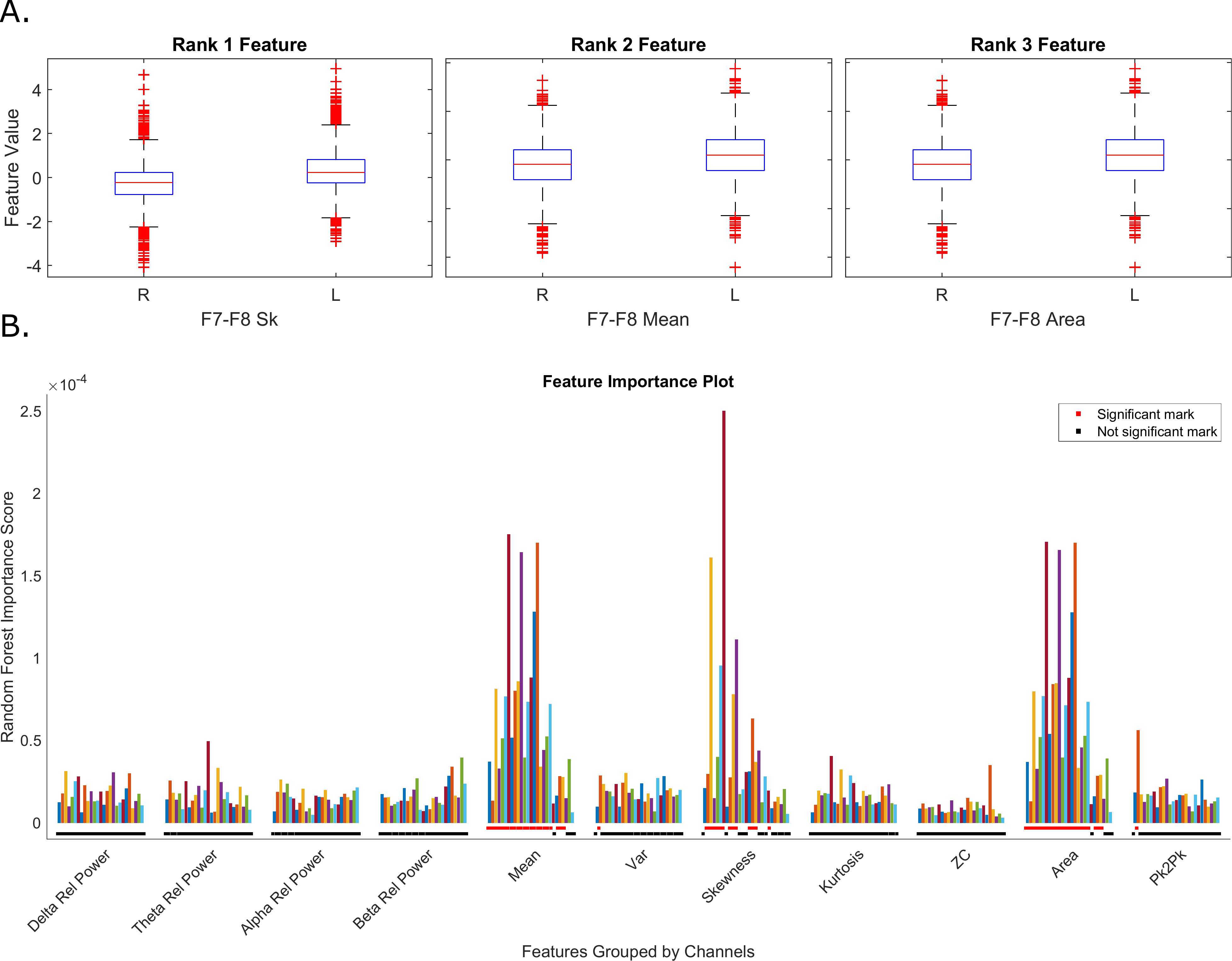}
    \caption{An overview of the extracted features is presented. (A) shows the box-plots of the top three features. (B) shows the feature importance scores measured by RF in addition to the significant features between the two classes, measured by non-parametric t-test.}
    \label{fig:5-5}
    \end{center}
\end{figure}

\textbf{Discussion:} 
Here, we analyze the distribution and contribution of the different features used in this study. In order to analyze the contribution of different features, we calculated the importance of each feature by employing Random Forest (RF) \cite{breiman2001random} for feature ranking. We did not use alternative feature importance measures such as Chi-Squared and F-measure \cite{forman2003extensive} since the features did not follow a normal distribution (one-sample Kolmogorov-Smirnov test: $p<\frac{0.05}{297}$). Figure \ref{fig:5-5}.A shows the three top features ranked using RF, namely skewness, mean, and area of F$7$-F$8$ sensor pair, where difference between the means of the two classes (L/R) can be observed. Further, Figure \ref{fig:5-5}.B illustrates the importance scores calculated using RF, along with the significance of each feature measured by non-parametric t-test at $p<\frac{0.05}{297}$. Finally, we select the top-$30$ significant subject-independent features and utilize them in the following paragraphs to explore the flow of information relevant to L/R hand movement through the sensors over time. The image in Table VII shows the sensor distribution based on the international $10$-$10$ system \cite{lotte2010exploring}. The table presents the ranking of the sensor pairs based on the number of features selected using the feature extraction and ranking method when the highest accuracy is achieved with our proposed solution. It can be observed that the FT$7$-FT$8$ sensor pair in the frontal-temporal lobe is the most dominant with $5$ features, followed by the T$9$-T$10$ sensor pair in the temporal lobe with $4$ features. The F$7$-F$8$ and T$7$-T$8$ sensor pairs, in the frontal and temporal lobes respectively, both have $3$ selected features, followed by F$5$-F$6$ with $2$, and the remaining sensor pairs with $1$ or $0$ features.

Next we analyze the flow of information at different times during the experiment. In Figure 6, the accuracy of the proposed method is depicted along with the standard deviation. The experiment started at $t = 0 s$ with the visual stimulus being presented to the subjects. As shown in the figure, in this stage, sensor pairs in the anterior-frontal (AF$3$-AF$4$ and AF$7$-AF$8$), parietal-occipital (PO$3$-PO$4$ and PO$7$-PO$8$), as well as occipital (O$1$-O$2$) lobes displayed the strongest features. This phenomenon is consistent with previous studies reporting that the visual cortex plays an important role in receiving and processing visual stimuli \cite{zeki1991direct, milner2008two}. The visual cortex occupies approximately $20\%$ space of the cerebral-cortex and is located in the occipital, parietal-posterior, and temporal lobes \cite{wandell2007visual}. Moreover, the information flow in the parietal-occipital and occipital regions were also consistent with the activities reported in \cite{wandell2007visual}.

\vspace{2mm}

The $P300$ wave is a type of Event-Related Potential (ERP) that is believed to be dominant in decision-making \cite{donchin2000mental}, and is usually within the range of $250 ms$ to $500 ms$ of the onset of visual stimulus \cite{yang2017usability}. Similar to the $P300$ wave, Simple Reaction Time (SRT) represents the delay between visual stimulus and response, during which the usage of the sensor pairs at $t = 0.25 s$ and $t = 0.50 s$ shows brain activity. This is consistent with \cite{woods2015factors}, reporting an average and standard deviation of $231 \pm 27 ms$ for SRT. During SRT, the previous highly informative sensor pairs in the parietal-occipital and occipital lobes gradually disappeared, and at $t = 0.25 s$ sensor pairs in the frontal-temporal (FT$7$-FT$8$) and temporal-parietal (TP$7$-TP$8$) lobes became more prevalent. The phenomenon is consistent with previous studies, stating that after the receipt of visual stimulus in the visual cortex, visual information is transferred through two disparate streams, notably ventral stream and dorsal stream. Ventral stream eventually reaches the temporal cortex, commonly known for image recognition \cite{goodale1992separate}. The visual stimulus is therefore processed to make the association between experiment instructions and performing L/R hand movement with the help of the relevant memory. Moreover, at $t = 0.25 s$ sensor pairs in the central-parietal (CP$3$-CP$4$ and CP$5$-CP$6$) lobes, as well as all the sensor pairs in the parietal lobe (P$1$-P$2$, P$3$-P$4$ and P$5$-P$6$) became more informative compared to $t = 0 s$. This activity is also consistent with the phenomenon reported in previous studies, stating that the dorsal stream eventually reaches the parietal-cortex, which contains action-relevant information \cite{goodale1992separate}. This is also supported by previous studies claiming that parietal lobe contributes predominantly to visual imagery and episodic memory \cite{pflugshaupt2014bottom}.

\begin{figure*}[t]
    \begin{center}
    \includegraphics[width=0.95\linewidth]{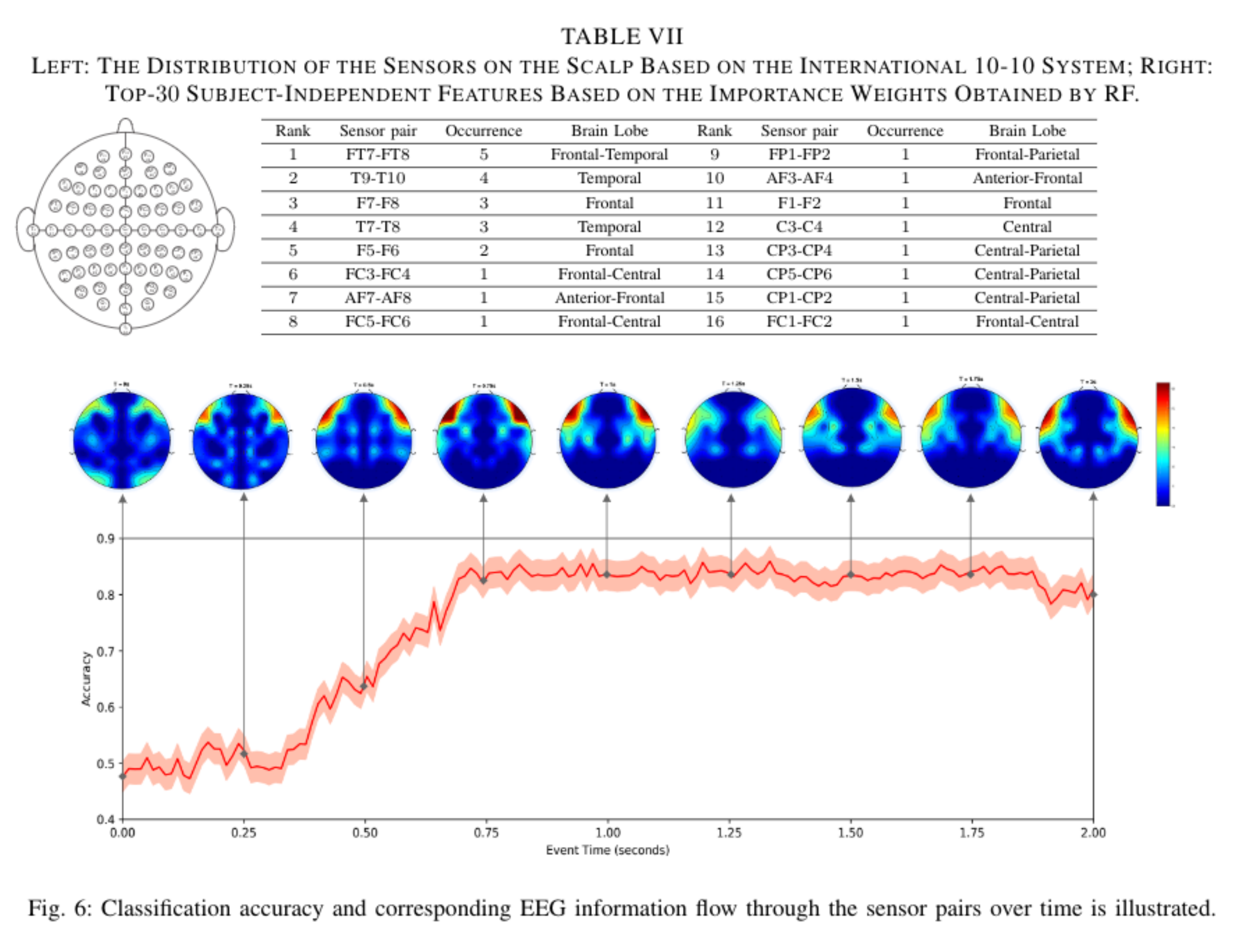} 
    \end{center}
\label{fig:6-6}
\end{figure*}

At $t = 0.5 s$, the active effect of sensor pairs in the parietal lobe diminished, which represents the completion of information flow in the dorsal stream. Consequently, sensor pairs in the central-parietal lobe (CP$1$-CP$2$, CP$3$-CP$4$ and CP$5$-CP$6$) became even more informative. This observed pattern is also consistent with previous related work claiming that the parietal lobe also contributes to hand and upper limb control and eye movements with visual information \cite{fogassi2005motor}.

At $t = 0.75 s$, sensor pairs in temporal (T$7$-T$8$ and T$9$-T$10$), frontal (F$7$-F$8$) and anterior-frontal (AF$7$-AF$8$) lobes reached their highest degree of use, thus having the highest discriminability for L/R hand movements. This phenomenon was consistent with the fact explained by previous studies that the frontal lobe contains pre-motor cortex and primary motor cortex (M$1$), where the pre-motor cortex first concatenates information from the parietal and frontal lobes, which is then delivered to M$1$. M$1$ is believed to be a generator of movement-specific commands \cite{dum2002motor}. Therefore, the classification rate reached its highest peak because of the completed information streams in occipital-temporal and occipital-parietal-frontal \cite{goodale1992separate}. Then, due to the high usage of sensor pairs in M$1$, the classification rate remained relatively stable until the movement ended \cite{dum2002motor}.

Lastly, from $t = 0.75 s$ to $t = 2.0 s$, the sensor pairs in the temporal and frontal lobes maintained the highest degree of usage, while those in the parietal lobe were much less frequently used. The other sensor pairs in the occipital lobe barely contributed to the classification task.

\section{Summary and Future Work}
A novel solution for classification of L/R hand movements from EEG signals was proposed. First the negative effects of signal artifacts was reduced, improving the quality of data. In the next step, a wide range of time and frequency domain features were exploited and used as inputs to an attention-based LSTM network. After studying the optimal LSTM hyper-parameter settings, extensive experiments were conducted with the EEG Movement Database over a large number of subjects ($103$). The performance evaluation with both intra-subject and cross-subject validation schemes showed very effective results with a high generalization capability, demonstrating the superiority of the proposed approach when compared to other benchmarking models and previous state-of-the-art methods. The robust performance achieved in this paper suggests that the proposed approach can be used in future research for a broad range of EEG-related classification tasks. Finally, a detailed analysis of the EEG information flow through the sensors over time is presented, reflecting the brain activity throughout the experiment.

Future work will focus on models capable of early detection or prediction of hand movements rather than classification using deep neural networks. Moreover, a possible limitation of our work is the use of hand-crafted features. As a result, in future work, feature extraction using CNNs will be explored, which may lead to a simpler and more robust solution. Lastly, to tackle the challenges in cross-subject classification, we will employ domain adaption techniques such as Wasserstein Generative Adversarial Network Domain Adaptation \cite{luo2018wgan} in order to distinguish and minimize the differences among subjects with adversarial training.

\section*{Acknowledgment}
The Titan XP GPU used for this research was donated by the NVIDIA Corporation.

\ifCLASSOPTIONcaptionsoff
\fi
\bibliographystyle{IEEEtran}
\bibliography{IEEEabrv,Bibliography}

\end{document}